\documentclass[12pt]{article}
\usepackage{amsmath,amssymb}
  \usepackage[T1]{fontenc}
  \usepackage[utf8]{inputenc}
\usepackage{lmodern}
\IfFileExists{upquote.sty}{\usepackage{upquote}}{}
\IfFileExists{microtype.sty}{
  \usepackage[]{microtype}
  \UseMicrotypeSet[protrusion]{basicmath} 
}{}
\usepackage{xcolor}
\usepackage[left=1.3in,right=1.3in,top=1in,bottom=1in]{geometry}
\providecommand{\tightlist}{%
  \setlength{\itemsep}{0pt}\setlength{\parskip}{0pt}}
\newlength{\cslhangindent}
\newlength{\csllabelwidth}
\newlength{\cslentryspacingunit} 
\newenvironment{CSLReferences}[2] 
 {
  \setlength{\parindent}{0pt}
  \ifodd #1
  \let\oldpar\par
  \def\par{\hangindent=\cslhangindent\oldpar}
  \fi
  \setlength{\parskip}{#2\cslentryspacingunit}
 }%
 {}
\usepackage{calc}

\usepackage{fancyhdr}
\usepackage{setspace}
\usepackage{amsmath, amsthm, amssymb, amsfonts}
\usepackage{graphicx}
\usepackage{bm}
\usepackage{bbm}
\def\tightlist{}
\usepackage{longtable}
\usepackage[format=plain,labelfont={bf},textfont=it]{caption}

\IfFileExists{bookmark.sty}{\usepackage{bookmark}}{\usepackage{hyperref}}
\IfFileExists{xurl.sty}{\usepackage{xurl}}{} 
\title{Synthetically generated text for supervised text analysis}
\author{Andrew Halterman\footnote{Assistant Professor, Michigan State
  University. ahalterman0@gmail.com}}
\date{February 17, 2023\footnote{First draft: 14 September 2022}}

\begin{document}
\maketitle
\begin{abstract}
Supervised text models are a valuable tool for political scientists but
present several obstacles to their use, including the expense of
hand-labeling documents, the difficulty of retrieving rare relevant
documents for annotation, and copyright and privacy concerns involved in
sharing annotated documents. This article proposes a partial solution to
these three issues, in the form of controlled generation of synthetic
text with large language models. I provide a conceptual overview of text
generation, guidance on when researchers should prefer different
techniques for generating synthetic text, a discussion of ethics, and a
simple technique for improving the quality of synthetic text. I
demonstrate the usefulness of synthetic text with three applications:
generating synthetic tweets describing the fighting in Ukraine,
synthetic news articles describing specified political events for
training an event detection system, and a multilingual corpus of
populist manifesto statements for training a sentence-level populism
classifier.
\end{abstract}

Word count: 9,582 (excluding Supplemental Information and table
contents)

\hypertarget{introduction}{%
\section{Introduction}\label{introduction}}

Supervised learning, in which a model is trained on labeled examples to
predict labels on new examples, is often the appropriate approach to
analyzing text when researchers want to categorize documents into known
categories or extract specific pieces of information from within a
document (Grimmer and Stewart 2013). Supervised text analysis brings
with it major costs, however: collecting human annotations on text is
potentially quite expensive, identifying rare event classes for
annotation can be difficult, and copyright, ethics, and privacy concerns
can make it difficult or impossible to share texts with annotators or
other researchers. These three issues limit the field's ability to build
supervised text analysis models and to evaluate previous work. Lowering
these costs would make it easier for political scientists to use
supervised models when they are the appropriate methodological tool for
their research.

This article proposes addressing these barriers by generating synthetic
text with desired content and style using large language models, a set
of computational models from natural language processing. Large language
models, such as GPT-2, GPT-3, and many others, are trained on large
corpus of text to predict the next word in a sequence of training data
(Radford et al. 2019). By using efficient neural net architectures,
models with a very large number of parameters, and large, diverse sets
of training data, language models can generate a sequence of tokens that
is likely to follow an input set of tokens.\footnote{The current state
  of the art for text generation uses transformer models (Vaswani et al.
  2018), which are an efficient architecture for training NLP models but
  a technical discussion of transformer models is outside the scope of
  this paper and the use of synthetic text does not depend on the
  specific neural network being used.} These large language models can
already generate news text or political text that is indistinguishable
from human-produced text (Zellers et al. 2019; Kreps, McCain, and
Brundage 2022).

This article describes how researchers in political science can use
large language models to lower the costs of supervised text analysis. It
provides guidance on the decisions that researchers face in generating
and using synthetic text, introduces a technique for measuring and
improving the quality of synthetic text, and discusses the serious
ethical pitfalls inherent in using synthetic text. It presents three
short applications from political science, illustrating how synthetic
text can be used in creating synthetic tweets describing the war in
Ukraine, synthetic news articles for event data detection models, and a
``zero shot'' (i.e., no labeled data) approach to building a
multilingual sentence-level populism classifier. It demonstrates that
synthetic data can be used to create tweets that are difficult to
distinguish from real tweets and that synthetic documents can be used to
train zero shot text classification models with no human annotation at
all. In general, a marginal labeled synthetic document does not improve
model performance as much as a marginal real document, meaning that
researchers face a tradeoff between accuracy and the benefits produced
by using synthetic text. However, the tradeoff can be reduced by using a
simple procedure I introduce for improving the quality of synthetically
generated text.

The synthetic data on populism contributes to a debate about the role
manifestos play in populist politics. I identify several manifestos
written by clearly populist parties that contain almost no populist
language. These results highlight the need to move beyond manifestos as
a primary source of data for studying populism, while also raising
theoretical questions about party decisionmaking around the manifesto
writing process.

\hypertarget{obstacles-to-using-supervised-text-analysis}{%
\subsection{Obstacles to using supervised text
analysis}\label{obstacles-to-using-supervised-text-analysis}}

Researchers conducting supervised text analysis face several obstacles.
The primary cost in developing a supervised model is \emph{labeling}, or
annotating text to serve as training data. Human annotations on
documents, such as document labels for classification or labeled
passages of text for information extraction, are expensive to collect.
Researchers need to define their concept of interest, create a codebook,
train annotators, pay them, and conduct quality assurance on the labels
they provide. Faced with the costs of obtaining labels, researchers may
turn to unsupervised techniques or avoid text analysis altogether.

A second obstacle is \emph{retrieval}, namely that annotators need to be
provided a set of text that has both relevant (positive class) documents
and non-relevant (negative class) documents. Because many of the
concepts of interest in political science are rare classes (Miller,
Linder, and Mebane 2020), a simple random sample from a corpus will
often not retrieve enough positive documents. For instance, if a
researcher is training a classifier to recognize police violence in news
text and it is only described in 0.5\% of news stories, a researcher
annotating a random sample of news stories would need to annotate 20,000
randomly sampled stories to expect to obtain 100 documents describing
police violence. To address the rare class issue, researchers will
over-sample relevant documents using keywords (e.g. Mueller and Rauh
2017), using active learning techniques (Miller, Linder, and Mebane
2020), or by exhaustively annotating an entire corpus (Halterman et al.
2021). Each of these techniques carries drawbacks in annotation cost,
low recall, or dependence on a model to suggest documents to label
(Halterman et al. 2021).

Finally, researchers face \emph{copyright} restrictions or
\emph{privacy} concerns that limit their ability to share annotated
documents. Most news articles are copyrighted, while other text, such as
social media posts, carry extra terms of service requirements that they
not be shared. Furthermore, even if researchers have the legal right to
share text, ethics and privacy concerns can preclude sharing documents
such as personal narratives, free-form survey responses, social media
posts containing personal information or discussions of sensitive
subjects. This makes it difficult or impossible to build on existing
annotated datasets or replicate existing methodological work. Generating
synthetic text in a controllable way, that is, with the ability to
direct the content and style of the text, can partially address these
issues

\hypertarget{using-language-models-to-generate-synthetic-text}{%
\section{Using language models to generate synthetic
text}\label{using-language-models-to-generate-synthetic-text}}

Generative language models learn to produce text by optimizing a
``language modeling'' objective: conditional on a sequence of tokens
(words), they predict which token is likely to follow.\footnote{For ease
  of explication, this paper focuses on the task of predicting the next
  token that follows a sequence of tokens, which is often referred to
  the computer science literature as a ``causal'' language model. Other
  language models, such as those in the BERT family (Jacob Devlin et al.
  2018), are ``bidirectional'', meaning they predict tokens to fill a
  gap in the middle of a sequence of tokens, conditioning on tokens that
  appear on either side of the missing word. This ``masked language
  modeling'' approach produces general purpose models that are well
  suited to a wide range of tasks, but generally do not perform as well
  as next word prediction models.} The parameters that control the
predicted probability of the next token in a sequence are learned
empirically from a large collection of training text. As an example, a
language model might learn that, given the input sequence ``the capital
of Germany is\_\_\_\_'\,', the token that has the highest probability of
coming next is ``Berlin''.

\hypertarget{formalizing-text-generation}{%
\subsection{Formalizing text
generation}\label{formalizing-text-generation}}

Formally, given a set of tokens \(W = \{w_1...w_n\}\), a language model
assumes that the probability of the sequence can be decomposed into the
probability of each token given the previous sequence of tokens:
\(p(W) = \prod_{i=1}^n p(w_i | w_{i-1},..., w_2, w_1\)).\footnote{A
  bidirectional model such as BERT predicts \(w_i\) in an \(n\) length
  sequence using tokens on either side:
  \(p(w_i | w_1, w_2,...,w_{i-1}, w_{i+1},...,w_n)\).} We can
approximate the conditional probability of the next token given the
previous tokens and trainable parameters \(\theta\):

\begin{equation}
\hat{p}(w_i) = f(w_{i-1}, w_{i-2},...,w_1, \theta).
\label{eq:p_token}
\end{equation}

To build intuition, we can consider a Markov chain, one of the simplest
techniques for generating a sequence of tokens. A Markov chain limits
the number of tokens used in predicting the following token
\(p(w_i | w_{i-1}, w_{i-2},...w_{k < n})\) and uses the raw empirical
frequency of each sequence in the training data as its parameters
\(\theta\). The advantages of newer language models are that they allow
longer sequences of text to inform their prediction of the next token
and their use of contextual word embeddings allow for more efficient
representations of words.

To generate text from a language model, we then sample a token
\(\hat{w_i}\) from the predicted distribution over the next word
\(\hat{p}(w_i)\) and a set of generation parameters \(\gamma\):

\begin{equation}
\hat{w}_i \sim \hat{p}(w_i), \gamma. 
\label{eq:gen}
\end{equation}

The generation parameters \(\gamma\) control how words are sampled from
the probability distribution over the next word. For instance,
\(\gamma\) might specify a ``greedy'' sampling strategy where the
highest probability word is always drawn, that words are sampled in
proportion to their predicted probability, or a more complicated process
that jointly generates several following tokens at once. While
\(\theta\), the parameters that govern the probability distribution over
words, are learned during training, \(\gamma\) can be varied later to
change how tokens are drawn. A later section of the paper provides a
technique for researchers to select optimal values of \(\gamma\) to
produce more realistic text.

\hypertarget{controlling-synthetic-text-generation}{%
\subsection{Controlling synthetic text
generation}\label{controlling-synthetic-text-generation}}

Thus, applied researchers who would like to influence which token
\(\hat{w}_i\) is produced next have three options: they can \emph{adapt}
the parameters \(\theta\) used to change the distribution
\(\hat{p}(w_i)\) given the previous tokens, they can \emph{prompt} by
changing the previous sequence of tokens (\(w_{i-1},w_{i-2},...\)), or
they can vary \(\gamma\) to change how the next token is sampled from
the distribution over the next token. These techniques are general,
working on current transformer-based neural networks, but also on older
technologies such as recurrent neural networks like LSTMs and on future
language models as
well.\footnote{Note that ChatGPT restricts users to prompting alone. This makes it simple to use but limits its flexibility.}

The \emph{adaptation} approach updates the weights \(\theta\) of a
pretrained model to affect the content or style of generated
text.\footnote{The natural language processing literature uses several
  terms to describe the process of updating a pretrained model's weights
  using new text, including ``fine tuning'', ``domain adaptation'', or
  ``additional pretraining''. Here, I use the term ``adaptation'' to
  refer to updating weights in a pretrained model to perform better on
  the language modeling task for a new corpus of text.} Researchers can
download generic pre-trained language models that have been trained on a
diverse set of text, including Wikipedia and unpublished novels (J.
Devlin et al. 2019), outbound links from Reddit (Radford et al. 2019),
or academic articles, crawled web pages, code repositories, movie
subtitles, and internet forums (e.g., Gao et al. 2020). Off-the-shelf
pretrained models may not reflect a researcher's desired style or
content, however, especially if it was not present in the original
pretraining data. Adapting an off-the-shelf model by providing it with
additional training data from a specific domain and updating the weights
\(\theta\) in the model to guide the text that the model produces,
changing \(p(\hat{w_i})\) as a result. An adaptation corpus should
reflect the type of synthetic text they are seeking to generate,
covering the full diversity of sources that they hope to reflect. When a
researchers use adaptation, they can employ a simple procedure
introduced below to maximize the similarity between the real and
synthetic text. The adaptation approach is used in the paper's first
application to generate synthetic tweets reporting battlefield updates
from the war in Ukraine.

Second, rather than updating the weights of a model, a research can
instead use \emph{prompting} to guide synthetic text generation. If a
researcher can provide the beginning of a document (\(w_{i-1},...w_n\)),
a large language model can generate a plausible continuation of the
document. For example, news articles can easily be prompted to contain
desired content using a hand-written headline that cues the content.
This approach is illustrated in the second application to generate news
stories describing armed conflict or violence by providing manually
written headlines to elicit stories with ``assault''-type events.

Recent off-the-shelf language models can generate text from abstract
prompts that describe the desired output, rather than simply continuing
from the starting tokens of some desired text (Liu et al. 2021). For
instance, GPT-3 can be provided with a prompt such as ``write a press
release in the style of a Republican House member'' and obtain a
plausible output without the need for a specific prompt or adapting a
language model on a corpus of Congressional Republican press releases.
This approach is useful when generating the desired content requires a
definition or explanation and for types of text that do not have a
natural summary--text format like news stories with their headlines.
Prompting with an explanation of the desired text is used in the third
application to generate populist party manifestos in 22 languages for 27
European countries.

When should researchers use adaptation and when should they prompt?
Table \ref{tab:when_to_use} provides an overview of how each technique
addresses the three challenges. Adaptation can provide them with a
version of their dataset that they can share freely, or can expand a
small set of labeled documents into a large corpus. Adaptation also
avoids the need for prompts--it is well known that language model output
is sensitive to changes in prompts (Zhao et al. 2021), potentially
producing text that does not match the context or style a researcher
wants. Adaptation may be necessary when a researcher's text is outside
the domain of the training data of the pretrained model. When adapting,
researchers can also quantitatively evaluate the quality of their
synthetic text by comparing it with the existing reference corpus.
Adaptation is usually more technically challenging, requiring
researchers to write code to update the pretrained model and often
requires greater computing resources, including access to a GPU.

Prompting has several other advantages over adaptation, beyond its
ability to work without an existing reference corpus. Prompts are
relatively transparent and can be published, allowing other researchers
can assess whether descriptive prompts accurately describe the concept
being prompted (e.g.~the definition of populism used in the third
application). Finally, prompts are easy to write and tweak to produce
synthetic text that matches the desired context (for example, in the
second application, to ensure that all aspects of the desired political
violence event class are included in the training corpus.)

The prompting approach is simple, but will fail if either the domain is
not covered well in the language model's training data (for example,
legal statues are not well represented in GPT-2's training data) or if
the desired text is difficult to prompt (for example, tweets lack the
convenient structure of news stories, where a headline usefully
summarizes the content of the story).

\begin{table}
\renewcommand{\arraystretch}{1.5}
\centering
\begin{tabular}{p{1in}p{1.5in}p{1.5in}p{1.5in}}
        & \textbf{Labeling} & \textbf{Retrieval} & \textbf{Copyright}\\
    \textbf{Adaptation}\\$p(w_i | w_{i-1}..., \textcolor{red}{\theta})$ & Useful: can expand an existing labeled corpus but requires existing labeled data. & Less useful: requires an existing corpus of the rare class, which is usually not available in a retrieval problem. However, useful for expanding an existing set of documents. & Ideal: can generate an uncopyrighted corpus that closely matches the original. Similarity can be quantified and improved. \\
    \bigskip

    \bigskip
    \textbf{Prompting}\\ $p(w_i | \textcolor{red}{w_{i-1}...}, \theta$ & Useful: can sometimes generate text with labels that are good enough for a zero-shot classifier. & Ideal: can generate a training corpus with a higher proportion of positive classes for labeling. & Useful: prompted text may not match the style/content as well as adapted model text. \\
    
\end{tabular}   
\caption{Matching the researcher's needs (labeling, retrieval, copyright) with the two techniques for guiding synthetic text generation (adaptation and prompting).}
\label{tab:when_to_use}
\end{table}

\hypertarget{improving-synthetic-text-quality-with-an-adversarial-classifier}{%
\subsection{Improving synthetic text quality with an adversarial
classifier}\label{improving-synthetic-text-quality-with-an-adversarial-classifier}}

Language models have a set of parameters \(\gamma\) that affect how a
word is sampled from \(\hat{p}(w)\). For GPT-2 and GPT-3, these include
the ``temperature'', ``top K'', and ``top P'', which control whether to
sample a high-probability next token (leading to simple, repetitive
text) or favor low probability next tokens (leading to more creative but
potentially nonsensical text).\footnote{The full details of GPT's
  generation parameters are beyond the scope of this paper. See Platen
  (2020).} Varying these hyperparameters greatly affects the quality of
the generated text, but little theoretical guidance exists on how to
select generation hyperparameters (Fu et al. 2021).\footnote{Recall from
  Eq. \ref{eq:gen} that the generation hyperparameters \(\gamma\)
  control the generation of text from an existing, trained language
  model. A separate set of parameters \(\theta\) are learned during
  training and affect the predicted probability of a following token
  (see Eq. \ref{eq:p_token}). Because large language models are
  expensive and time consuming to train from scratch, a discussion of
  the hyperparameters that affect how language models are trained is
  beyond the scope of this article. For more details on language model
  training see Dodge et al. (2020).}

I introduce a simple ``adversarial'' procedure for selecting the best
generation hyperparameters for generating text, drawing on the intuition
that the harder it is for a classifier to distinguish between real and
synthetic text, the higher the quality of synthetic text. The procedure
is adversarial in the sense that the worse the classifier performs, the
better the quality of the synthetic text. For each set of
hyperparameters, a researcher generates \(n\) synthetic documents and
samples \(n\) real documents from the existing corpus. They then split a
classifier on a training set of both synthetic and real documents with
the objective of predicting whether the document is real or synthetic.
Evaluating the classifiers's performance on a test set provides a
quantitative measure of the synthetic text's quality. Decreasing model
performance indicates that real and synthetic documents are increasingly
difficult to distinguish. The set of hyperparameters that results in the
lowest classification accuracy is thus the set that generates the best
documents.

Alternatively, a researcher could assess the synthetic document quality
by fitting a structural topic model (Roberts et al. 2013) with the
document's real/synthetic label as a covariate and examining the
difference in topics. This allows a research to check whether the
synthetic text covers the entire domain of real text or if certain
topics are differentially represented between the corpora, at the cost
of slightly greater complexity.

\hypertarget{end-to-end-synthetic-text-pipeline}{%
\subsection{End-to-end synthetic text
pipeline}\label{end-to-end-synthetic-text-pipeline}}

\begin{figure}
\center
\includegraphics[width=\columnwidth]{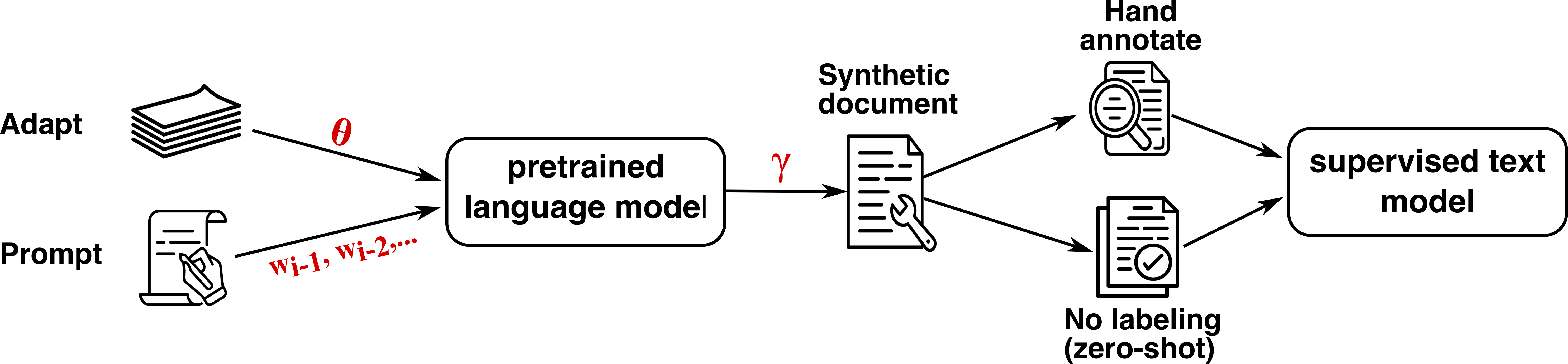}
\caption{Overview of the synthetic text process for supervised learning. Researchers can affect the content and style of synthetic documents by changing language model parameters ($\theta$), by providing new prompts ($w_{i-1}, w_{i-2},...$), or the sampling parameters ($\gamma$). Researchers then decide whether to hand-label the resulting synthetic documents or whether to use them as-is to train a zero-shot model, i.e, one without any hand-labeled data.}
\label{fig:pipeline}
\end{figure}

After they generate text, researchers then have two options for how to
use it. Figure \ref{fig:pipeline} shows a high-level overview of the
entire process. First, they can treat the text as unlabeled and collect
annotations on it in the same way they would with real text. By using
controlled synthetic text, they have addressed the copyright or usage
restrictions of real text and can share their annotated text freely, and
have hopefully addressed the retrieval problem as well. This approach is
used in the first and second applications below. A second option exists
when researchers are conducting document classification and believe that
their prompting strategy reliably generates documents with the desired
class label. In this situation, researchers can train their document
classifier directly on the synthetic text without hand labeling.

\hypertarget{ethics}{%
\subsection{Ethics}\label{ethics}}

Generating synthetic text presents serious ethical concerns. Synthetic
text can include factual errors, conspiracy theories, or offensive
statements. To avoid any possibility of synthetic text being mistaken
for real text, researchers working with synthetic text should always
attach a disclaimer directly to any synthetic text any time it is saved
or stored, clearly indicating that the text is synthetic.\footnote{For
  example,
  \texttt{\textless{}!-\/-SYNTHETIC\ TEXT!\ Do\ not\ trust\ the\ factual\ content\ of\ this\ text.\ Generated\ by\ \textless{}author\textgreater{},\ \textless{}email\textgreater{}\ to\ train\ a\ populist\ speech\ classifier.\ -\/-\textgreater{}}}\\
The disclaimer should only be removed temporarily as a final step before
fitting a model to avoid the possibility of synthetic data being
mistaken for real text. Annotators should be briefed on the use of
synthetic text and the annotation interface should clearly state that
they are working with synthetic text, which likely contains factual
errors. Any synthetic text reported in published work must be clearly
marked (e.g.~\textsf{\footnotesize [SYNTH]}) and the accompanying text
clearly explain its use and potential bias.

While synthetic text may be difficult to distinguish from real text in
its style or writing and thus useful for training a model to recognize
certain linguistic features, its factual content will be imaginary and
thus completely unsuitable for answering substantive questions on its
own.

Third, it is well known that language models learn the biases present in
their training data (Caliskan, Bryson, and Narayanan 2017). Researchers
should validate that the models that they train on synthetic text are
not relying on group stereotypes when making predictions. Curating the
text using to adapt the model or using prompts that break the
association between groups and stereotyped traits offers a partial
solution, but greater research into the prevalence and mitigation of
these harms is needed.

Finally, if researchers are using synthetic text as a privacy-preserving
strategy as some researchers have suggested (Ororbia II, Linder, and
Snoke 2018; Li et al. 2021), they should opt to label their text rather
than using zero-shot techniques to ensure that none of their synthetic
text includes mentions of real people or other sensitive information
that may have been learned by the adapted language model.

\hypertarget{applications}{%
\section{Applications}\label{applications}}

The following section presents three short applications that illustrate
the decisions researchers face in generating synthetic text, including
when to adapt and when to prompt. The first application, on generating
synthetic tweets describing the war in Ukraine, illustrates an
adaptation approach and introduces a simple method for improving the
quality of the generated text. The second and third applications use two
approaches to prompting to generate synthetic news stories describing
political violence and to create data for training a sentence-level
populism classifier. The third application also highlights limitations
in studying populism using manifestos.

\hypertarget{adapting-language-models-for-copyright-free-tweets-identifying-weapons-in-the-ukraine-war}{%
\subsection{Adapting Language Models for Copyright-Free Tweets:
Identifying Weapons in the Ukraine
War}\label{adapting-language-models-for-copyright-free-tweets-identifying-weapons-in-the-ukraine-war}}

Social media posts are a major source of text for political scientists
but platform terms of use and privacy concerns greatly limit
researchers' ability to share or publish their collected posts. This a
concern in situations where researchers are collecting expensive
annotations on posts to serve as training data or making methodological
claims about new methods that require the original training data to
replicate. Transparency and reproducibility will both be greatly helped
by releasing the raw training data used to train supervised learning
models on social media posts. In the case of Twitter, researchers
generally provide tweets to other researchers in ``dehydrated'' form,
consisting of tweet IDs. Other researchers then ``rehydrate'' the tweets
by querying the Twitter API for the original tweet and merge with the
original annotation. If, in the meantime, the original tweet has been
deleted, the researcher will not be able to obtain the tweet. While the
rate of deletion is generally below 5\%, some politically significant
situations, such as tweets related to Brexit, can have rates of deletion
as high as 33\% (Bastos 2021).

As political scientists begin to collect more data about the ongoing war
in Ukraine (Zhukov 2022), supervised models trained on tweets will
provide an important source of information. This application shows that
synthetic tweets can be generated that are difficult to distinguish from
real tweets. A named entity recognition system trained on synthetic
tweets reaches the same accuracy as one trained on real tweets, but
requires approximately 50\% more annotations to reach the same
performance.

I collect a set of around 20,000 real tweets from four Twitter accounts
that report detailed information on the fighting in Ukraine.\footnote{Specifically,
  @uaweapons, @osinttechnical, @oryxspioenkop, and @markito0171.}
Because the synthetic tweets should closely match the actual tweets and
because tweets are more difficult to prompt than news articles, which
have a convenient headline--body structure, I opt for a adaptation
approach to text generation. I adapt a large language model,
specifically GPT-2, on this set of tweets to produce a language model
that is well suited to generating tweets about the war (@ Wolf et al.
2020). By adapting the model, I can both ensure that the generated text
is similar to real tweets about the war in Ukraine, as well as
eliminating the need to provide a specific prompt to generate text.
Adaptation is especially valuable for this application because the
generic training data for GPT-2 was collected before the war, making it
impossible to produce accurate synthetic tweets about the war without
adapting the model on tweets written after February 2022.

To improve the quality of the synthetic tweets, I apply the adversarial
method introduced above. Across 56 combinations of hyperparameters, the
classification accuracy varies from 0.98 to a low of 0.64, indicating a
large effect of hyperparameters of tweet generation.\footnote{See SI
  Figure \ref{fig:tweet_gen_parameters}.} Table \ref{tab:ukraine_tweets}
reports a random sample of synthetic tweets using the best
hyperparameters (that is, the ones producing the lowest accuracy for the
discriminator model).

\begin{table}
\begin{tabular}{p{0.9\textwidth}}
\hline
 \textsf{[SYNTH]} The system is relatively good at engaging low/medium armored targets, like \textbf{BTRs}, \textbf{MT-LBs}, \textbf{APCs} and \textbf{SPGs} \\
\textsf{[SYNTH]}  I think people got the wrong impression from today’s press conference, where Lukashenko said “I do not fear Western military threats but Russia is prepared to pay a heavy price for any military action. \\
\textsf{[SYNTH]}  This is mostly because air defence is weak, and even non \textbf{TB2s} could get shot down. Only a very few aircraft flew today, with the majority of them from the western part of Ukraine. In the north of Ukraine a lack of \textbf{TB2s} has caused large losses. The Ukrainians are probably using the drones to spot artillery strikes. \\
\textsf{[SYNTH]}  Tanks on the other side of the Irpin River \\
\hline
\end{tabular}
\caption{Selected synthetically generated tweets from a random sample of 10
generated from a GPT-2 model adapted/fine tuned on 20,000 tweets reporting open source
intelligence on the war in Ukraine. Weapon annotations shown in bold. See the SI for the full list of 10
randomly selected tweets. Due to Twitter's restrictions on including 
tweets in published work, no actual tweets are shown.  GPT-2
generation parameters ($\gamma$): top\_p= 0.90, top\_k= 50, temperature= 1.5}
\label{tab:ukraine_tweets}
\end{table}

\hypertarget{comparing-performance-of-real-and-synthetic-tweets}{%
\subsubsection{Comparing performance of real and synthetic
tweets}\label{comparing-performance-of-real-and-synthetic-tweets}}

\begin{figure}
\centering
\includegraphics{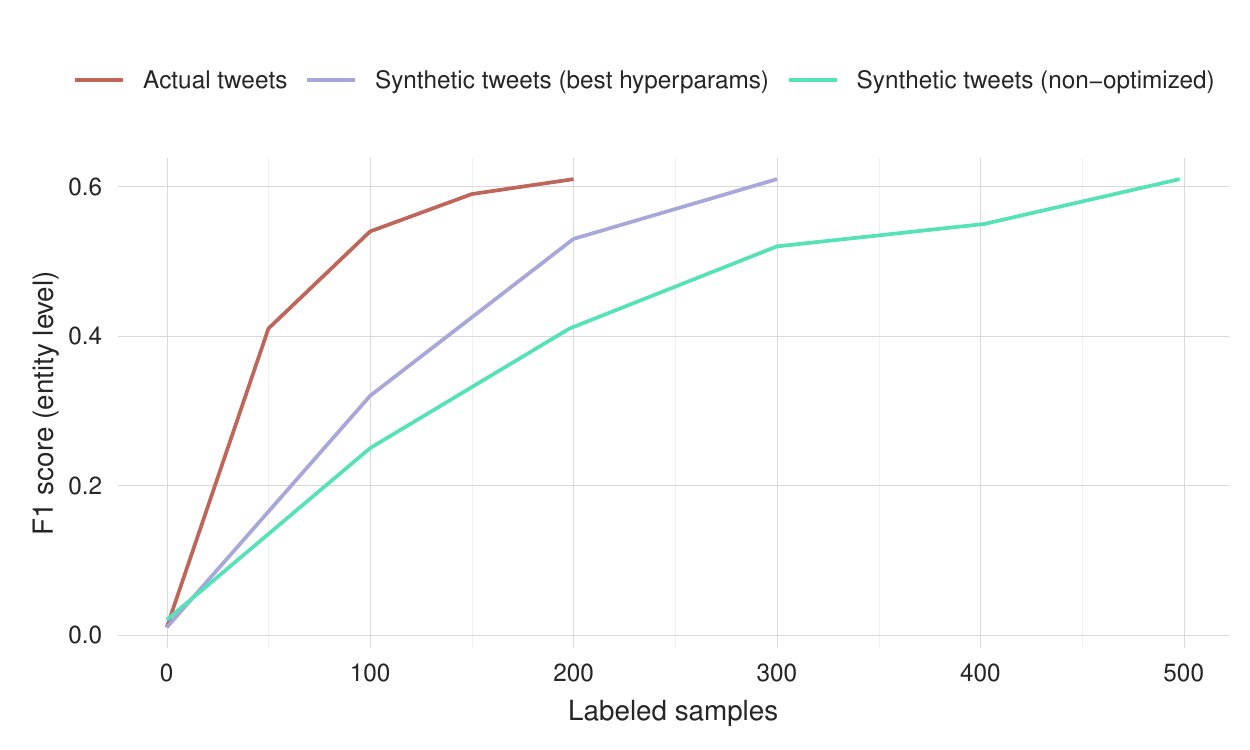}
\caption{Test set performance of a named entity recognition model detecting a
    \textsc{weapon} class, trained on annotated actual tweets and annotated
    synthetic tweets. A model trained on annotated synthetic tweets generated from
    the optimal hyperparameters found using the adversarial technique requires
    around 300 examples to reach the performance of a model trained on 200 annotated
    actual tweets. In contrast, when using non-optimized synthetic tweets, 500
    labeled synthetic examples are required to reach the performance of 200 actual
    tweets.}
\label{fig:tweets}
\end{figure}

How well do these synthetic tweets work in practice? I hand annotate
1,600 tweets with span-level labels on the specific weapons systems
described in the tweets. The annotated set includes 200 real tweets, 600
synthetic tweets generated from non-optimized parameters, 600 tweets
generated using the parameters selected by the adversarial tuning
method, with an additional 200 real tweets as evaluation data. I train a
named entity recognition (NER) model to identify mentions of specific
weapons in the text.\footnote{I use spaCy 3.1.2's small
  \texttt{en\_core\_web\_sm} model as a base and the default training
  values set by Prodigy (Honnibal and Montani 2017; Montani and Honnibal
  2018). Better absolute performance could be achieved with a larger
  model, but I expect the relative performance to be the same.} Figure
\ref{fig:tweets} reports the accuracy for the model trained on actual
and synthetic tweets at different training set sizes and evaluated on
labeled actual tweets.\footnote{More specifically, I use span-based F1
  score as an accuracy measure, where precision is the proportion of
  identified named entities that are correct, recall is the proportion
  of named entities identified by the model, and F1 is the harmonic mean
  of the two: A per-token F1 score would be higher, but the relevant
  metric here is the relative performance from the two data sources, not
  the absolute performance of the NER model.} The figure shows that the
performance of the NER model improves as it has access to more labeled
training examples across actual tweets, synthetic tweets generated
without any generation parameter tuning, and synthetic tweets generated
with optimal parameters. A marginal non-optimized synthetic tweet is
significantly less valuable than a real labeled tweet: the model
requires 500 non-optimized synthetic training examples to reach the
performance it can attain with 200 labeled actual tweets. Applying the
adversarial technique to select the optimal generation hyperparameters
reduces the gap significantly: only 300 synthetic tweets are required to
reach the same performance as the actual tweet model, rather than 500
without optimizing.

While using synthetic tweets carries a cost, namely that they are
somewhat less useful than a marginal actual tweet in training a
classifier, it provides a major benefit in that the labeled training
data can be published without violating the Twitter terms of service.
Publishing the complete training data is especially important for
methodological work, where replicating a model's performance and
allowing others to build on previous datasets is a key component of
progress.

\hypertarget{application-2-generating-rare-documents-for-human-labelingtraining-an-event-data-classifier}{%
\subsection{Application 2: Generating rare documents for human
labeling--training an event data
classifier}\label{application-2-generating-rare-documents-for-human-labelingtraining-an-event-data-classifier}}

Event data is a major source of quantitative information for researchers
in international relations and comparative politics (Beieler et al.
2016). Automated event data systems increasingly rely on machine
learning models to identify events, but despite decades of investment,
no corpus of news text with comprehensive political event labels
exists.\footnote{Some datasets with limited coverage exist,
  e.g.~annotation on actions taken by police in India (Halterman et al.
  2021). Event datasets in computer science/computational linguistics
  such as ACE (Doddington et al. 2004) have event definitions that are
  largely not relevant for political science and very expensive to
  license.} The lack of a gold-standard set of labeled news articles
makes it difficult to evaluate event classification models or to develop
new classifiers as machine learning techniques improve. This application
shows how the three obstacles to supervised text analysis can be
overcome with \emph{prompting}. The adaptation approach used in the
Ukraine war tweets example is not suited to this application, because
adaptation requires an existing set of articles with a known event type,
which we do not have. News text is well-represented in the training
corpora for many language models, making them suitable to generating
synthetic news articles.

A simple way to generate synthetic news articles with desired content is
write a headline that reflects the event type or concept we would like
to have a story about. For example, to generate stories about
disinformation or information operations, we can write the headline
``Foreign `information operation' spreading disinformation uncovered''.
We can provide this headline to a large language model, in this case
GPT-2 (large) (Radford et al. 2019) and generate a synthetic news story
prompted by the title. We can increase the diversity of the training
text by modifying the byline in the prompt to refer to different news
sources and cities.

\begin{quote}
\textsf{[PROMPT]}

\emph{Foreign `information operation' spreading disinformation
uncovered}

\emph{BELGRADE (Reuters)}

\textsf{[-- SYNTHETIC STORY --]} In an unusual development this week,
Serbian President Tomislav Nikolic called the foreign media a threat to
the nation's security.

``Foreign `information operations' against us and our country (are) the
ones spreading the disinformation against us,'' Nikolic said in
{[}\ldots{]}

\textsf{[-- SYNTHETIC STORY --]}
\end{quote}

If we keep the same headline but change the byline to ``BRUSSELS (local
sources)'', we can generate a story about misinformation occurring in a
completely different context (although note that the story itself is
misinformation):

\begin{quote}
\textsf{[PROMPT]}

\emph{Foreign `information operation' spreading disinformation
uncovered}

\emph{BRUSSELS (local sources)}

\textsf{[-- SYNTHETIC STORY --]}

A report by Belgian daily La Presse has shown that the United States is
planning on deploying military `information operations' in Europe, which
could be used for `hybrid war' scenarios. The report said that the
Pentagon is sending US forces to Western Europe to create `fake news'
and create a `vast digital surveillance network' on political parties,
activists, media commentators and'subversives' with {[}\ldots{]}

\textsf{[-- SYNTHETIC STORY --]}
\end{quote}

\begin{figure}
\includegraphics{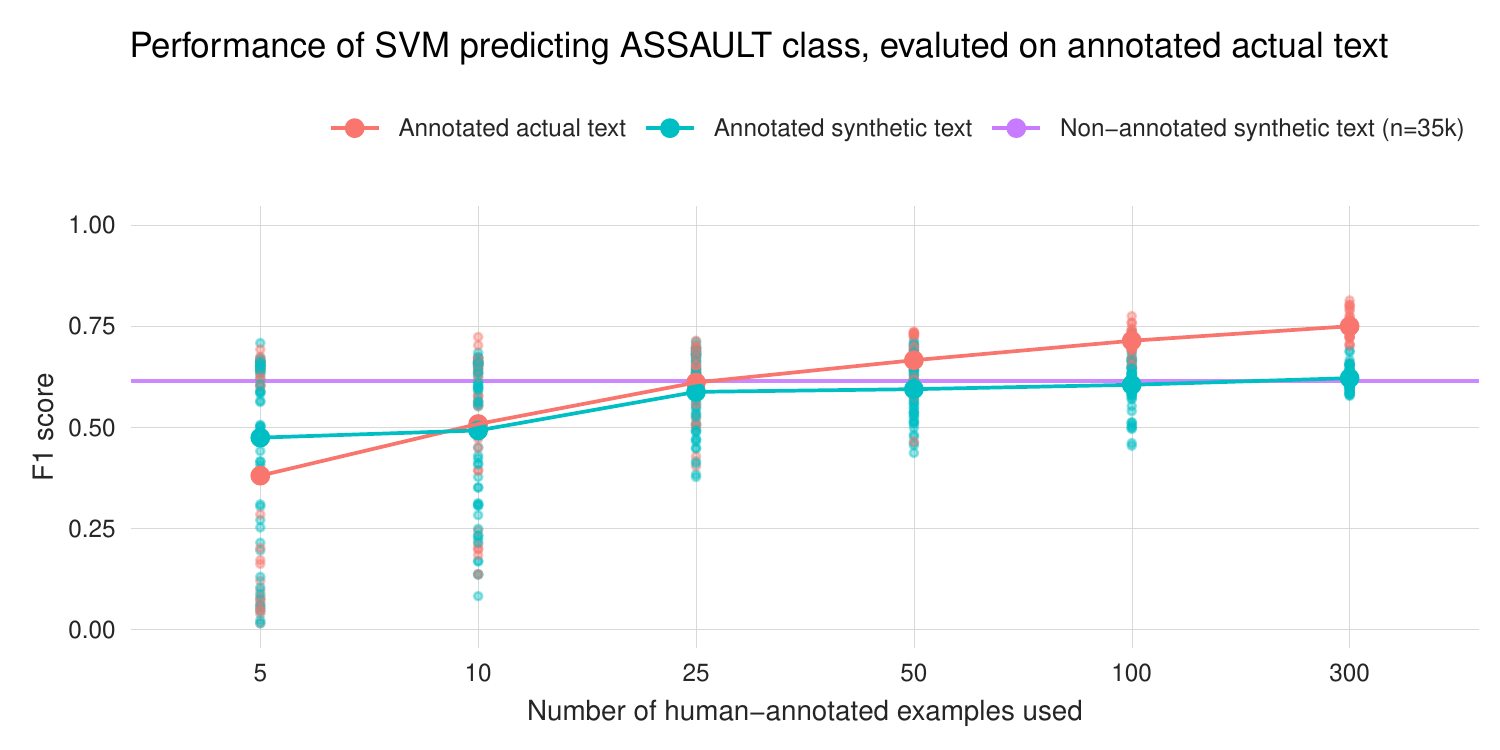}
\caption{Performance of SVM classifier predicting the \textsc{assault} class
using three sets of training data: annotated real news articles, annotated
synthetic articles, and non-annotated synthetic articles (no human labels). Using synthetic
documents incurs an accuracy cost, but only after around 50 annotated articles.
Using 35,000 non-annotated synthetic articles (i.e., assuming that every article
includes the desired event type) performs at least as well as labeled synthetic
stories, offering a zero-shot classification option. Smaller points indicate 25
random train/test splits, lines show mean performance.}
\label{fig:event_data}
\end{figure}

(\textbf{halterman2022plover?}) introduce a new dataset of hand
annotations on 12,952 news stories for 16 defined event classes,
including an \textsf{ASSAULT} event class that includes a broad range of
political violence and armed conflict, which is used as a training and
evaluation set for this application.\footnote{According to the dataset's
  event ontology, ``ASSAULT events are deliberate actions which can
  potentially result in substantial physical harm'', including military
  assaults, attacks, kidnappings, terrorist attacks, ethnic cleansing,
  torture, beatings, etc.} Figure \ref{fig:event_data} compares the
performance of classifiers trained on three sets of data:
human-annotated actual news stories, human-annotated synthetic stories,
and unlabeled synthetic stories. In the last case, synthetic documents
generated with headlines meant to prompt \textsf{ASSAULT} are assumed to
do so, and stories generated with non-\textsf{ASSAULT} headlines are
assumed to not contain \textsf{ASSAULT} events.\footnote{See SI A for
  details on the headlines used and the total number of synthetic
  stories.}

Figure \ref{fig:event_data} shows that a marginal labeled real document
improves out-of-sample classifier performance more than an additional
labeled synthetic document, indicating that researchers face a tradeoff
between accuracy and the retrieval/copyright benefits of synthetic
text.\footnote{I use an SVM without tuning the hyperparameters for
  simplicity and computational ease. Better tuning or more sophisticated
  models would outperform the SVM, but the interest here is in the
  relative, not absolute performance.} The result also show, however,
that a model trained exclusively on unannotated synthetic documents
performs at least as well as one that uses annotated synthetic
documents. Because synthetic documents are essentially free to produce,
a large number can be generated and used to train a ``zero shot''
classifier that performs well for the task.

Thus, synthetic text can partially address all three obstacles to
building a supervised text classifier for identifying political events
in text. It addresses the \emph{retrieval} problem that researchers face
in collecting annotations on rare classes. Even if a text generation
process is not completely accurate in generating articles with the
desired class, it still produces a corpus with a much greater
concentration of relevant documents than a random sample from the corpus
would generate. In this case, it also partially solves the
\emph{labeling} problem. If a researcher is willing to accept a
classifier with somewhat worse performance, they can forgo an annotation
process entirely and use only the labels from their prompts. Regardless
of if they hand annotate or take the synthetic labels as-is, they will
now have a labeled dataset that they can freely share, improving
reproducibility and allowing other researchers to develop improved
classifiers with their data.

\hypertarget{application-3-synthetic-data-for-zero-shot-classifierstraining-a-sentence-level-populist-classifier}{%
\subsection{Application 3: Synthetic data for zero-shot
classifiers--training a sentence-level populist
classifier}\label{application-3-synthetic-data-for-zero-shot-classifierstraining-a-sentence-level-populist-classifier}}

As attention to populist parties has grown, so too has the
methodological work on identifying populism in text, including in party
manifestos (Rooduijn and Pauwels 2011; Hawkins et al. 2019; Di Cocco and
Monechi 2021; Dai and Kustov 2022; Jankowski and Huber 2023; Breyer
2022). A key challenge has been to identify populism in short text, such
as a sentence or a paragraph, in order to estimate the degree or amount
of populism in a document. Given that no dataset exists that labels
populist speech at the sentence level, recent work has proposed training
sentence-level supervised classifiers to identify populism using
manifesto-level labels (Di Cocco and Monechi 2021). This approach has
been criticized, for, among other things, for relying on document-level
labels to train a sentence-level classifier when most sentences in a
populist party's manifesto will not be recognizably populist (Jankowski
and Huber 2023). This application illustrates a new method for
identifying populism at the sentence level across 27 European countries
in 22 languages. I use a \emph{prompting} approach to generate synthetic
populist manifesto statements, and then train a classifier on the
synthetic data to identify populist statements in real manifesto text. I
find evidence that the classifier reliably identifies populist rhetoric
in text and also identify a major limitation in studying populism using
manifesto text. Several populist parties have virtually no measured
populist rhetoric in their manifestos, which is bourne out through
careful reading of the manifestos. This finding has implications for the
study of populism and highlights the inherent limitations of using
manifesto text to assess parties' populism.

I employ a prompting approach to generate populist manifesto statements,
given that adaptation requires a labeled dataset and no hand-labeled
dataset of populism in manifesto sentences has been published. Political
manifestos are much rarer than news stories, making it difficult for
smaller language models or those trained on less diverse text to
accurately generate manifesto text. Moreover, manifestos cannot be
easily prompted with headlines in the same way that news stories are.
Larger language models such as GPT-3 (Brown et al. 2020) can use more
abstract prompts than smaller models like GPT-2, including definitions
or descriptions of the desired text. Including a description of the
desired text allows researchers to incorporate a abstract explanations
of political concepts to generate relevant text and allows other
researchers to examine the definitions employed in the prompting.

\hypertarget{measuring-populism}{%
\subsubsection{Measuring populism}\label{measuring-populism}}

I employ a conceptualization of populism drawing on Mudde's (2004)
``thin'' definition of populism, which focuses on its rhetorical aspects
and worldview rather than on specific policy positions. Mudde defines
populism as ``an ideology that considers society to be ultimately
separated into two homogeneous and antagonistic groups, `the pure
people' versus `the corrupt elite', and which argues that politics
should be an expression of the volonté générale (general will) of the
people'' (2004, 543). Both Rooduijn (2019) and Hunger and Paxton (2022)
caution against conflating populism with right-wing populism, the
``radical right,'' nativism, or anti-political establishment parties. I
attempt to provide conceptually clear examples of populist rhetoric by
writing prompts that are meant to elicit populist statements from the
language model, without conflating populism with other political
stances, such as anti-migrant, right-wing, or Eurosceptic positions, or
oppostion to the party in power.\footnote{I thank Michael Jankowski for
  helpful comments on this point.} Table \ref{tab:populism_prompts}
shows the two prompts I use to generate populist statements. One prompt
is a ``thin'' prompt that restates Mudde's definition of populism, and
the other includes non-politician elites, which are an important but
sometimes overlooked component of populism (Jungkunz, Fahey, and Hino
2021). Because GPT-3 was trained on a multilingual corpus, we can
specify the desired country and language in the prompt to obtain
non-English text, even with an English language prompt. By inserting
each country and its associated language(s) into the prompt and varying
the sampling hyperparemeters (\(\gamma\)), I generate a set of 5,357
synthetic populist sentences.

\begin{table}[h]
\renewcommand{\arraystretch}{1.5}
\begin{tabular}{p{0.3\textwidth}p{0.7\textwidth}}
Description &  Prompt   \\
  \hline
   Populist (thin~definition) & Populist rhetoric sees politics as a conflict with good, common, or "real" people on one side, and out-of-touch, evil, or self-serving elites on the other.\newline Write ten statements that a populist party in \{country\} might make (in \{language\}):\\
   & \textit{example output} [\textsf{SYNTH}] \textit{``We're committed to giving voice to those who have been ignored or left behind by mainstream politics."}\\
 Populist (style prompt) & A populist party in \{country\} believes that politics is corrupted by self-interested elites, unelected bureaucrats, croynism, and big business. It wants to take power back for ordinary people. Write 12 statements that a \{country\_adjective\} populist party might make (in the \{language\} language) in the style of a political manifesto:\\
 & \textit{example output} [\textsf{SYNTH}] \textit{``We believe that the people of Ireland are sovereign, and that the government should be accountable to them."}
\end{tabular}
\caption{Prompts used to generate populist text.}
\label{tab:populism_prompts}
\end{table}

To train a populism classifier, I also require non-populist sentences,
which I generate using two kinds of prompts. First, I draw policy
position descriptions from the Manifesto Project (Volkens et al. 2021).
The Manifesto Project hand labels sentences or phrases in political
manifestos for their policy positions. I take the ten most frequently
identified codes in their dataset and use the policy positions'
descriptions from their codebook as prompts\footnote{See SI Table
  \ref{tab:cmp_prompts} examples.} Jungkunz, Fahey, and Hino (2021)
caution that measures of populism often pick up on opposition to the
current ruling party, as opposed to populism \emph{per se}. To mitigate
this issue, I add an additional set of ten hand-written non-populist
prompts to cue criticism of other political parties and dissatisfaction
with current policies, which were lacking in the Manifesto Project
prompted text.\footnote{See SI Table \ref{tab:non_populist_hand_prompts}
  for the ten prompts.} By once again varying the countries, languages,
and generation hyperparameters used in the prompts, I obtain 36,509
non-populist synthetic sentences. Note that neither the populist nor
non-populist text is generated with party names in the prompts,
mitigating the risk of the model picking up party names as a predictive
feature (Jankowski and Huber 2023).

I then train a supervised text classifier on the synthetic sentences. I
fine-tune a multilingual sentence transformer model (Reimers and
Gurevych 2020) and classifier on the sentences using SetFit, an
efficient model for short text classification (Tunstall et al.
2022).\footnote{note that in this context, ``fine tuning'' refers to
  updating the weights of the model on a downstream classification task,
  not to ``adaptation''.} In training the model, I assume that the
statements generated with the populist prompt are indeed examples of
populist rhetoric, and that non-populist prompts generate non-populist
rhetoric. A thorough evaluation of the model's performance is difficult,
given the lack of a published dataset of labeled populist sentences. As
a first evaluation, however, the model achieves an accuracy of 0.93 and
macro F1 score of 0.85 in identifying populist statements in synthetic
validation data. This indicates that the classifier can reliably
distinguish between sentences from populist and non-populist prompts,
but does not necessarily generalize to the model's ability to identify
populist statements in real text.

Next, I move from synthetic text to real text taken from party
manifestos. I use the Manifesto Project's dataset of party manifesto
text. I apply the newly trained populism classifier to the each of the
sentences or phrases in the Manifesto Project corpus (Volkens et al.
2021), producing a predicted \([0, 1]\) populism score for each
sentence. Examining the scored sentences by hand provides some further
validity for the method. Given all manifesto sentences from the United
Kingdom Independence Party, a populist party, the model identifies the
three sentences as having the highest populism scores:

\begin{itemize}
  \setlength\itemsep{0em}
\item ``Politics is corrupted by self-interest and big business."
\item ``These professional politicians don't want us to run our own country or control our own lives."
\item ``An unaccountable elite revels in mutual back-scratching and cronyism."\footnote{See SI Table \ref{tab:ukip_random} for 10 randomly selected UKIP sentences and their associated populism scores.}
\end{itemize}

\hypertarget{populism-in-populist-party-manifestos}{%
\subsubsection{Populism in Populist Party
Manifestos}\label{populism-in-populist-party-manifestos}}

\begin{figure}[ht]
\begin{center}
\includegraphics{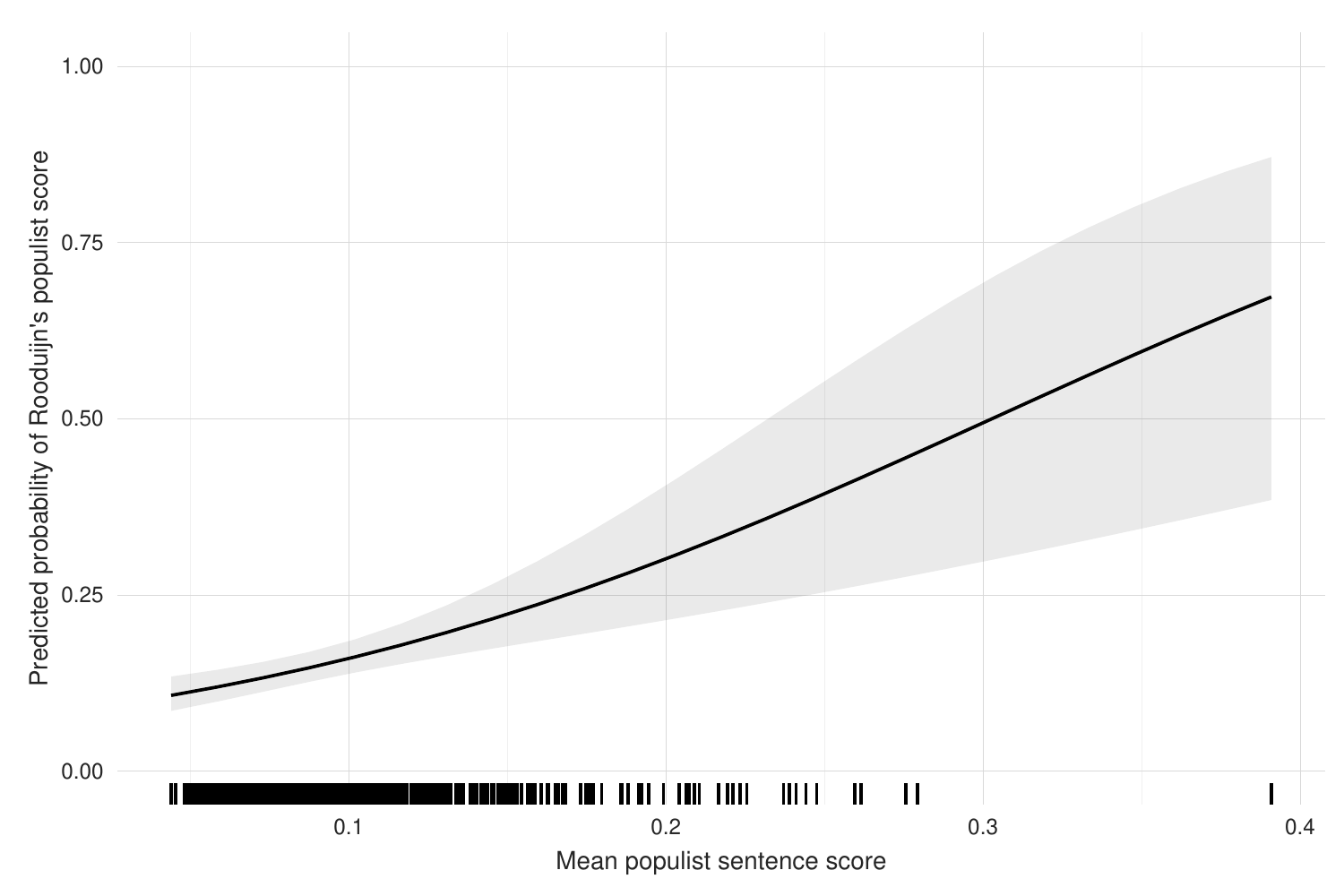}
\end{center}
\caption{Manifesto-derived populist scores are postivitely and significantly correlated with Rooduijn's populist party measure. Predictions from a logistic regression of Rooduijn's populist party measure on manifesto-derived populism scores.}
\label{fig:pop_score_regression}
\end{figure}

While populism is inherently an anti-elite ideology, political party
manifestos are primarily intended for elite consumption rather than for
the general public (Dolezal et al. 2012; Eder, Jenny, and Müller 2017;
Harmel 2018). Manifestos are thus not the most natural place for
populist rhetoric to appear. Comparing the degree of populism in party
manifestos to the party's PopuList score (Rooduijn et al. 2019) shows
that populist parties do include more populist rhetoric in their
manifestos than non-populist parties (Figure
\ref{fig:pop_score_regression}) The average populism score of sentences
in a party's manifestos is a positive and significant predictor of the
party's PopuList score.\footnote{The regression is a logistic regression
  of the PopuList score on the average populism score of sentences in
  the party's manifesto. The regression is significant at the 0.0001
  level.} This suggests that populist parties do include more populist
rhetoric in their manifestos than non-populist parties.

However, the degree of populism in a party's manifesto is not a good
predictor of the party's PopuList score: the area under the receiver
operator curve (AUC) for the logitistic regression is 0.63,\footnote{An
  intuitive understanding of the AUC is the probability that the
  manifesto text model produces a higher prediction for a randomly
  selected positive case (here, a populist party) than a randomly
  selected negative case would.}. What can we learn from the ``false
negatives'' from our classifier? Table \ref{tab:non_populist} lists the
parties that are coded as populist in the PopuList dataset that do not
have any sentences with a predicted populism score above 0.5.

\begin{table}[t]
\centering
\begin{tabular}{rlllr}
  \hline
  Country & Party & Election Date & Mean Populism Score \\ 
  \hline
  Italy & People of Freedom & 2013-02-24 & 0.05 \\ 
  Italy & Go Italy & 2018-03-04 & 0.05 \\ 
  Italy & Brothers of Italy & 2018-03-04 & 0.06 \\ 
  Italy & Northern League & 2013-02-24 & 0.05 \\ 
  Greece & Independent Greeks & 2015-01-25 & 0.10 \\ 
  Germany & Party of Democratic Socialism & 1994-10-16 & 0.06 \\ 
  Latvia & Zatlers' Reform Party & 2011-09-17 & 0.08 \\ 
  Poland & Self-Defence of the Polish Republic & 2001-09-23 & 0.06 \\ 
  Romania & People's Party - Dan Dianconescu & 2012-12-09 & 0.06 \\ 
  Slovakia & Alliance of the New Citizen & 2002-09-20 & 0.05 \\ 
   \hline
\end{tabular}
\caption{Party-elections rated as populist by PopuList [@rooduijn2019populist] whose manifestos include no sentences with a predicted populism score above 0.5. Average populism score shown in the second column.}
\label{tab:non_populist}
\end{table}

I randomly select the Northern League's 2013 manifesto for hand coding.
While some of the sentences include some statements that are
populism-adjacent, including opposition to bank bailouts, or right wing,
such as support for the traditional family, none contain overt
statements of hostility to elites or support for ``common people''
against elites.\footnote{See Table \ref{tab:northern_league_hand} for
  details on the hand coding.}

Future substantive work should attempt to provide a better theoretical
account of why populist parties make different decisions about how much
populist rhetoric to include in their manifestos. Populist parties
moderate their manifestos in expectation of future coalition membership
(Harmel 2018) or engage in a ``campaign gamble'', increasing their use
of populism when their electoral chances are lower (Dai and Kustov 2022)
or when in opposition (Breyer 2022). Future methodological work could
extend work on populist rhetoric in campaign speeches and materials
{[}e.g., Hawkins et al. (2019); dai2022politicians{]} to measure the
discrepencies between these materials and manifestos.

\hypertarget{conclusion}{%
\section{Conclusion}\label{conclusion}}

Three main obstacles to supervised text analysis in political science,
the problems of labeling, retrieval, and copyright, can be addressed in
part by generating synthetic text with the content or style that a
researcher desires. Different applications will call for different
approaches to generating synthetic text, including using off-the-shelf
language models, adapted language models, or using very large models
that can be prompted with directions about the desired text. A new
adversarial technique helps researchers select the optimal
hyperparameters to generate synthetic text that is difficult to
distinguish from real text. Each of these approaches is illustrated in
three short applications from political science, demonstrating that
synthetic text can address the retrieval and copyright issues, and
sometimes the problem of labeling, albeit with some penalty in accuracy.

A researcher might wonder whether the step of generating synthetic text
is needed at all. If a large language model can reliably generate text
with a desired label, it should also be able to apply that label
directly to actual text (Ornstein, Blasingame, and Truscott 2022). While
models are likely to improve greatly in the near future, the synthetic
text approach has several benefits over a zero shot classifier approach.
First, many of the best models, such as GPT-3, are hosted by third
parties that require payment for their use. While still cheaper than
hand-coding a large corpus of text (Ornstein, Blasingame, and Truscott
2022), paying to obtain annotations on a large corpus of text can
quickly become expensive. Second, hosted models change rapidly, raising
difficulties for future researchers in replicating earlier work.
Finally, and most significantly, the classification outputs of language
models are often opaque and difficult to evaluate. By using large
language models to generate synthetic text and using more well
understood models for classification, including bag-of-words models,
researchers can evaluate the quality of the generated text and employ
classifiers that are faster to run and easier to understand.

This approach to generating synthetic text is applicable to a wide range
of tasks. Future work can explore the use of synthetic text to evaluate
pre-analysis plans for analyzing free-form text in survey responses, to
allow greater transparency in interviews or field notes while preserving
privacy, and in developing improved techniques for guiding the content
and quality of the synthetic text. Researchers can also explore hybrid
approaches that use a mixture of real and synthetic text. Existing
labeled text can be used to generate more synthetic text, using the
existing text to prompt or adapt a model, and then augment the original
training corpus with synthetic examples to increase its size and
diversity.

Researchers in natural language processing are developing more
sophisticated techniques for controlled text generation to allow
covariates to guide text generation or ensure the factuality of the
generated text (e.g. Dathathri et al. 2020; Prabhumoye, Black, and
Salakhutdinov 2020; Yogatama, Masson d'Autume, and Kong 2021). Once they
mature, these methods will offer additional options for applied
researchers to generate controlled text. However, these methods will
still operate either by modifying the conditioning sequence of text
(\(w_{i-1}...w_1\)), modifying the language model's parameters
(\(\theta\)), or changing how words are sampled from the probability
distribution (\(\gamma\)). Even as the technology to generate synethetic
text improves, applied researchers in political science will still face
the same decisions about when and how to generate synthetic text and how
to obtain labels for their synthetic text.

\hypertarget{acknowledgements}{%
\subsection{Acknowledgements}\label{acknowledgements}}

Thank you to Benjamin Bagozzi, Jill Irvine, Michael Jankowski, Adam
Lauretig, Erin Rossiter, and Brandon Stewart for helpful comments on
earlier drafts. This work was first presented at PolMeth 2022 at
Washington University in St.~Louis and at New Directions in Text as Data
2022 at Cornell Tech. Thanks also to the MIT SuperCloud and Lincoln
Laboratory Supercomputing Center for providing the computing resources
(Reuther et al. 2018).

This work grew out of two ongoing collaborations. The event
classification application uses hand-annotated data from a project with
Benjamin Bagozzi, Phil Schrodt, Andy Beger, and Grace Scarborough. The
populism application draws on ongoing substantive work with Shahryar
Minhas, Christian Houle, and Nicolas Bichay.

Portions of this paper were sponsored by the Political Instability Task
Force (PITF). The PITF is funded by the Central Intelligence Agency. The
views expressed in this paper are the author's alone and do not
represent the views of the US Government.

\hypertarget{references}{%
\subsection{References}\label{references}}

\hypertarget{refs}{}
\begin{CSLReferences}{1}{0}
\leavevmode\vadjust pre{\hypertarget{ref-bastos2021account}{}}%
Bastos, Marco. 2021. {``This Account Doesn't Exist: Tweet Decay and the
Politics of Deletion in the Brexit Debate.''} \emph{American Behavioral
Scientist} 65 (5): 757--73.

\leavevmode\vadjust pre{\hypertarget{ref-beieler2016generating}{}}%
Beieler, John, Patrick T Brandt, Andrew Halterman, Erin Simpson, and
Philip A Schrodt. 2016. {``Generating Political Event Data in Near Real
Time: Opportunities and Challenges.''} In \emph{Computational Social
Science}, edited by R. Michael Alvarez. Cambridge University Press.

\leavevmode\vadjust pre{\hypertarget{ref-breyer2022populist}{}}%
Breyer, Magdalena. 2022. {``Populist Positions in Party Competition: Do
Parties Strategically Vary Their Degree of Populism in Reaction to Vote
and Office Loss?''} \emph{Party Politics}, 13540688221097082.

\leavevmode\vadjust pre{\hypertarget{ref-brown2020language}{}}%
Brown, Tom B, Benjamin Mann, Nick Ryder, Melanie Subbiah, Jared Kaplan,
Prafulla Dhariwal, Arvind Neelakantan, et al. 2020. {``Language Models
Are Few-Shot Learners.''} \emph{Advances in Neural Information
Processing Systems} 33 (1877--1901).

\leavevmode\vadjust pre{\hypertarget{ref-caliskan2017semantics}{}}%
Caliskan, Aylin, Joanna J Bryson, and Arvind Narayanan. 2017.
{``Semantics Derived Automatically from Language Corpora Contain
Human-Like Biases.''} \emph{Science} 356 (6334): 183--86.

\leavevmode\vadjust pre{\hypertarget{ref-dai2022politicians}{}}%
Dai, Yaoyao, and Alexander Kustov. 2022. {``When Do Politicians Use
Populist Rhetoric? Populism as a Campaign Gamble.''} \emph{Political
Communication}, 1--22.
https://doi.org/\url{https://doi.org/10.1080/10584609.2022.2025505}.

\leavevmode\vadjust pre{\hypertarget{ref-dathathri2019plug}{}}%
Dathathri, Sumanth, Andrea Madotto, Janice Lan, Jane Hung, Eric Frank,
Piero Molino, Jason Yosinski, and Rosanne Liu. 2020. {``Plug and Play
Language Models: A Simple Approach to Controlled Text Generation.''}
\emph{ICLR}.

\leavevmode\vadjust pre{\hypertarget{ref-devlin2018bert}{}}%
Devlin, Jacob, Ming-Wei Chang, Kenton Lee, and Kristina Toutanova. 2018.
{``{BERT}: Pre-Training of Deep Bidirectional Transformers for Language
Understanding.''} \emph{arXiv Preprint arXiv:1810.04805}.

\leavevmode\vadjust pre{\hypertarget{ref-devlin2019bert}{}}%
Devlin, J., Ming-Wei Chang, Kenton Lee, and Kristina Toutanova. 2019.
{``{BERT}: Pre-Training of Deep Bidirectional Transformers for Language
Understanding.''} In \emph{NAACL-HLT}.

\leavevmode\vadjust pre{\hypertarget{ref-di2021populist}{}}%
Di Cocco, Jessica, and Bernardo Monechi. 2021. {``How Populist Are
Parties? Measuring Degrees of Populism in Party Manifestos Using
Supervised Machine Learning.''} \emph{Political Analysis}, 1--17.

\leavevmode\vadjust pre{\hypertarget{ref-doddington2004automatic}{}}%
Doddington, George R, Alexis Mitchell, Mark A Przybocki, Lance A
Ramshaw, Stephanie Strassel, and Ralph M Weischedel. 2004. {``The
Automatic Content Extraction ({ACE}) Program-Tasks, Data, and
Evaluation.''} In \emph{LREC}, 2:1.

\leavevmode\vadjust pre{\hypertarget{ref-dodge2020fine}{}}%
Dodge, Jesse, Gabriel Ilharco, Roy Schwartz, Ali Farhadi, Hannaneh
Hajishirzi, and Noah Smith. 2020. {``Fine-Tuning Pretrained Language
Models: Weight Initializations, Data Orders, and Early Stopping.''}
\emph{arXiv Preprint arXiv:2002.06305}.

\leavevmode\vadjust pre{\hypertarget{ref-dolezal2012life}{}}%
Dolezal, Martin, Laurenz Ennser-Jedenastik, Wolfgang C Müller, and Anna
Katharina Winkler. 2012. {``The Life Cycle of Party Manifestos: The
Austrian Case.''} \emph{West European Politics} 35 (4): 869--95.

\leavevmode\vadjust pre{\hypertarget{ref-eder2017manifesto}{}}%
Eder, Nikolaus, Marcelo Jenny, and Wolfgang C Müller. 2017. {``Manifesto
Functions: How Party Candidates View and Use Their Party's Central
Policy Document.''} \emph{Electoral Studies} 45: 75--87.

\leavevmode\vadjust pre{\hypertarget{ref-fu2021theoretical}{}}%
Fu, Zihao, Wai Lam, Anthony Man-Cho So, and Bei Shi. 2021. {``A
Theoretical Analysis of the Repetition Problem in Text Generation.''} In
\emph{Proceedings of the AAAI Conference on Artificial Intelligence},
35:12848--56. 14.

\leavevmode\vadjust pre{\hypertarget{ref-gao2020pile}{}}%
Gao, Leo, Stella Biderman, Sid Black, Laurence Golding, Travis Hoppe,
Charles Foster, Jason Phang, et al. 2020. {``The {P}ile: An 800{GB}
Dataset of Diverse Text for Language Modeling.''} \emph{arXiv Preprint
arXiv:2101.00027}.

\leavevmode\vadjust pre{\hypertarget{ref-grimmer2013text}{}}%
Grimmer, Justin, and Brandon M Stewart. 2013. {``Text as Data: The
Promise and Pitfalls of Automatic Content Analysis Methods for Political
Texts.''} \emph{Political Analysis} 21 (3): 267--97.
\url{https://doi.org/10.1093/pan/mps028}.

\leavevmode\vadjust pre{\hypertarget{ref-halterman2021corpus}{}}%
Halterman, Andrew, Katherine A Keith, Sheikh Muhammad Sarwar, and
Brendan O'Connor. 2021. {``Corpus-Level Evaluation for Event {QA}: The
{I}ndia{P}olice{E}vents Corpus Covering the 2002 {G}ujarat Violence.''}
\emph{Findings of the Association for Computational Linguistics}.

\leavevmode\vadjust pre{\hypertarget{ref-harmel2018hows}{}}%
Harmel, Robert. 2018. {``The How's and Why's of Party Manifestos: Some
Guidance for a Cross-National Research Agenda.''} \emph{Party Politics}
24 (3): 229--39.

\leavevmode\vadjust pre{\hypertarget{ref-hawkins2019measuring}{}}%
Hawkins, Kirk A, Rosario Aguilar, Bruno Castanho Silva, Erin K Jenne,
Bojana Kocijan, and Cristóbal Rovira Kaltwasser. 2019. {``Measuring
Populist Discourse: The Global Populism Database.''} In \emph{EPSA
Annual Conference in Belfast, UK, June}, 20--22.

\leavevmode\vadjust pre{\hypertarget{ref-honnibal2017spacy}{}}%
Honnibal, Matthew, and Ines Montani. 2017. {``spaCy 2: Natural Language
Understanding with Bloom Embeddings, Convolutional Neural Networks and
Incremental Parsing.''} \emph{To Appear}.

\leavevmode\vadjust pre{\hypertarget{ref-hunger2022buzzword}{}}%
Hunger, Sophia, and Fred Paxton. 2022. {``What's in a Buzzword? A
Systematic Review of the State of Populism Research in Political
Science.''} \emph{Political Science Research and Methods} 10 (3):
617--33.

\leavevmode\vadjust pre{\hypertarget{ref-jankowski2022correlation}{}}%
Jankowski, Michael, and Robert A Huber. 2023. {``When Correlation Is Not
Enough: Validating Populism Scores from Supervised Machine-Learning
Models.''} \emph{Political Analysis}, no. 1--15.
\url{https://doi.org/doi:10.1017/pan.2022.32}.

\leavevmode\vadjust pre{\hypertarget{ref-jungkunz2021populist}{}}%
Jungkunz, Sebastian, Robert A Fahey, and Airo Hino. 2021. {``How
Populist Attitudes Scales Fail to Capture Support for Populists in
Power.''} \emph{Plos One} 16 (12): e0261658.

\leavevmode\vadjust pre{\hypertarget{ref-kreps2022all}{}}%
Kreps, Sarah, R Miles McCain, and Miles Brundage. 2022. {``All the News
That's Fit to Fabricate: AI-Generated Text as a Tool of Media
Misinformation.''} \emph{Journal of Experimental Political Science} 9
(1): 104--17.

\leavevmode\vadjust pre{\hypertarget{ref-Li2021AreSC}{}}%
Li, Jianfu, Yujia Zhou, Xiaoqian Jiang, Karthik Natarajan, Serguei V. S.
Pakhomov, Hongfang Liu, and Hua Xu. 2021. {``Are Synthetic Clinical
Notes Useful for Real Natural Language Processing Tasks: A Case Study on
Clinical Entity Recognition.''} \emph{Journal of the American Medical
Informatics Association : JAMIA}.

\leavevmode\vadjust pre{\hypertarget{ref-liu2021pre}{}}%
Liu, Pengfei, Weizhe Yuan, Jinlan Fu, Zhengbao Jiang, Hiroaki Hayashi,
and Graham Neubig. 2021. {``Pre-Train, Prompt, and Predict: A Systematic
Survey of Prompting Methods in Natural Language Processing.''}
\emph{arXiv Preprint arXiv:2107.13586}.

\leavevmode\vadjust pre{\hypertarget{ref-miller2019active}{}}%
Miller, Blake, Fridolin Linder, and Walter R Mebane. 2020. {``Active
Learning Approaches for Labeling Text: Review and Assessment of the
Performance of Active Learning Approaches.''} \emph{Political Analysis},
1--20.

\leavevmode\vadjust pre{\hypertarget{ref-montani2018prodigy}{}}%
Montani, Ines, and Matthew Honnibal. 2018. {``Prodigy: A New Annotation
Tool for Radically Efficient Machine Teaching.''} \emph{Artificial
Intelligence} to appear.

\leavevmode\vadjust pre{\hypertarget{ref-mudde2004populist}{}}%
Mudde, Cas. 2004. {``The Populist Zeitgeist.''} \emph{Government and
Opposition} 39 (4): 542--63.

\leavevmode\vadjust pre{\hypertarget{ref-mueller2017reading}{}}%
Mueller, Hannes, and Christopher Rauh. 2017. {``Reading Between the
Lines: Prediction of Political Violence Using Newspaper Text.''}
\emph{American Political Science Review}, 1--18.

\leavevmode\vadjust pre{\hypertarget{ref-ornstein_blasingame_truscott_2022_parrot}{}}%
Ornstein, Joseph T., Blasingame Elise N., and Jake S. Truscott. 2022.
{``How to Train Your Stochastic Parrot: Deep Language Models for
Political Texts.''} \emph{PolMeth Conference Paper}.

\leavevmode\vadjust pre{\hypertarget{ref-ororbia2018using}{}}%
Ororbia II, Alexander G, Fridolin Linder, and Joshua Snoke. 2018.
{``Using Neural Generative Models to Release Synthetic {T}witter Corpora
with Reduced Stylometric Identifiability of Users.''} \emph{arXiv
Preprint arXiv:1606.01151}.

\leavevmode\vadjust pre{\hypertarget{ref-scikit-learn2011}{}}%
Pedregosa, F., G. Varoquaux, A. Gramfort, V. Michel, B. Thirion, O.
Grisel, M. Blondel, et al. 2011. {``Scikit-Learn: Machine Learning in
{P}ython.''} \emph{Journal of Machine Learning Research} 12: 2825--30.

\leavevmode\vadjust pre{\hypertarget{ref-von2020generate}{}}%
Platen, Patrick von. 2020. {``How to Generate Text: Using Different
Decoding Methods for Language Generation with Transformers.''}
\emph{Hugging Face Blog}.

\leavevmode\vadjust pre{\hypertarget{ref-prabhumoye2020exploring}{}}%
Prabhumoye, Shrimai, Alan W Black, and Ruslan Salakhutdinov. 2020.
{``Exploring Controllable Text Generation Techniques.''} \emph{arXiv
Preprint arXiv:2005.01822}.

\leavevmode\vadjust pre{\hypertarget{ref-radford2019language}{}}%
Radford, Alec, Jeffrey Wu, Rewon Child, David Luan, Dario Amodei, and
Ilya Sutskever. 2019. {``Language Models Are Unsupervised Multitask
Learners,''} 9.

\leavevmode\vadjust pre{\hypertarget{ref-reimers-2020-multilingual-sentence-bert}{}}%
Reimers, Nils, and Iryna Gurevych. 2020. {``Making Monolingual Sentence
Embeddings Multilingual Using Knowledge Distillation.''} In
\emph{Proceedings of the 2020 Conference on Empirical Methods in Natural
Language Processing}. Association for Computational Linguistics.
\url{https://arxiv.org/abs/2004.09813}.

\leavevmode\vadjust pre{\hypertarget{ref-reuther2018interactive}{}}%
Reuther, Albert, Jeremy Kepner, Chansup Byun, Siddharth Samsi, William
Arcand, David Bestor, Bill Bergeron, et al. 2018. {``Interactive
Supercomputing on 40,000 Cores for Machine Learning and Data
Analysis.''} In \emph{2018 {IEEE} High Performance Extreme Computing
Conference ({HPEC})}, 1--6. IEEE.

\leavevmode\vadjust pre{\hypertarget{ref-roberts2013structural}{}}%
Roberts, Margaret E, Brandon M Stewart, Dustin Tingley, Edoardo M
Airoldi, et al. 2013. {``The Structural Topic Model and Applied Social
Science.''} In \emph{Advances in Neural Information Processing Systems
Workshop on Topic Models: Computation, Application, and Evaluation}.

\leavevmode\vadjust pre{\hypertarget{ref-rooduijn2019state}{}}%
Rooduijn, Matthijs. 2019. {``State of the Field: How to Study Populism
and Adjacent Topics? A Plea for Both More and Less Focus.''}
\emph{European Journal of Political Research} 58 (1): 362--72.

\leavevmode\vadjust pre{\hypertarget{ref-rooduijn2011measuring}{}}%
Rooduijn, Matthijs, and Teun Pauwels. 2011. {``Measuring Populism:
Comparing Two Methods of Content Analysis.''} \emph{West European
Politics} 34 (6): 1272--83.

\leavevmode\vadjust pre{\hypertarget{ref-rooduijn2019populist}{}}%
Rooduijn, Matthijs, Stijn Van Kessel, Caterina Froio, Andrea Pirro,
Sarah De Lange, Daphne Halikiopoulou, Paul Lewis, Cas Mudde, and Paul
Taggart. 2019. {``The PopuList: An Overview of Populist, Far Right, Far
Left and Eurosceptic Parties in Europe.''}

\leavevmode\vadjust pre{\hypertarget{ref-tunstall2022efficient}{}}%
Tunstall, Lewis, Nils Reimers, Unso Eun Seo Jo, Luke Bates, Daniel
Korat, Moshe Wasserblat, and Oren Pereg. 2022. {``Efficient Few-Shot
Learning Without Prompts.''} \emph{arXiv Preprint arXiv:2209.11055}.

\leavevmode\vadjust pre{\hypertarget{ref-vaswani2018tensor2tensor}{}}%
Vaswani, Ashish, Samy Bengio, Eugene Brevdo, Francois Chollet, Aidan N
Gomez, Stephan Gouws, Llion Jones, et al. 2018. {``Tensor2tensor for
Neural Machine Translation.''} \emph{arXiv Preprint arXiv:1803.07416}.

\leavevmode\vadjust pre{\hypertarget{ref-volkens_etal2021manifesto}{}}%
Volkens, Andrea, Tobias Burst, Werner Krause, Pola Lehmann, Nicolas AND
Regel Matthieß Theres AND Merz, Bernhard Weels, and Lisa Zehnter. 2021.
{``The Manifesto Data Collection. Manifesto Project
({MRG/CMP/MARPOR}).''} \emph{Version 2021a, Berlin: Wissenschaftszentrum
Berlin f{ü}r Sozialforschung (WZB)}.
https://doi.org/\url{https://doi.org/10.25522/manifesto.mpds.2021a}.

\leavevmode\vadjust pre{\hypertarget{ref-wolf2020transformers}{}}%
Wolf, Thomas, Julien Chaumond, Lysandre Debut, Victor Sanh, Clement
Delangue, Anthony Moi, Pierric Cistac, et al. 2020. {``Transformers:
State-of-the-Art Natural Language Processing.''} In \emph{Proceedings of
the 2020 Conference on Empirical Methods in Natural Language Processing:
System Demonstrations}, 38--45.

\leavevmode\vadjust pre{\hypertarget{ref-yogatama2021adaptive}{}}%
Yogatama, Dani, Cyprien de Masson d'Autume, and Lingpeng Kong. 2021.
{``Adaptive Semiparametric Language Models.''} \emph{Transactions of the
Association for Computational Linguistics} 9: 362--73.

\leavevmode\vadjust pre{\hypertarget{ref-zellers2019grover}{}}%
Zellers, Rowan, Ari Holtzman, Hannah Rashkin, Yonatan Bisk, Ali Farhadi,
Franziska Roesner, and Yejin Choi. 2019. {``Defending Against Neural
Fake News.''} In \emph{Advances in Neural Information Processing Systems
32}.

\leavevmode\vadjust pre{\hypertarget{ref-zhao2021calibrate}{}}%
Zhao, Zihao, Eric Wallace, Shi Feng, Dan Klein, and Sameer Singh. 2021.
{``Calibrate Before Use: Improving Few-Shot Performance of Language
Models.''} In \emph{International Conference on Machine Learning},
12697--706. PMLR.

\leavevmode\vadjust pre{\hypertarget{ref-zhukov2022viina}{}}%
Zhukov, Yuri M. 2022. {``VIINA: Violent Incident Information from News
Articles on the 2022 Russian Invasion of Ukraine.''} In \emph{Ann Arbor:
University of Michigan, Center for Political Studies}.
https://github.com/zhukovyuri/VIINA.

\end{CSLReferences}

Image credit for Figure \ref{fig:pipeline}: Flaticon.com

\newpage

\appendix

\hypertarget{supplemental-information}{%
\section{Supplemental Information}\label{supplemental-information}}

I estimate 40 hours of GPU usage for this paper. Using energy
consumption data on the GPU I use and standard New England electricity
carbon intensity, this would produce around 3kg of CO\(_2\) emissions,
corresponding to around 7 miles driven by an average US passenger car.
https://mlco2.github.io/impact/\#compute.{]}

\hypertarget{supplemental-information-a-selecting-generation-hyperparameters-for-synthetic-tweets}{%
\subsection{Supplemental Information A: Selecting generation
hyperparameters for synthetic
tweets}\label{supplemental-information-a-selecting-generation-hyperparameters-for-synthetic-tweets}}

\renewcommand{\thefigure}{A\arabic{figure}}
\renewcommand{\thetable}{A\arabic{table}}
\setcounter{figure}{0}
\setcounter{table}{0}

I consider 56 combinations of generation hyperparameters and generate
1,000 synthetic tweets for each set.

\begin{itemize}
\tightlist
\item
  epoch \(\in\) \{1, 3\}
\item
  top\_p \(\in\) \{0.8, 0.90, 0.95, 0.99\}
\item
  temperature \(\in\) \{0.3, 0.5, 0.7, 1, 1.3, 1.5, 1.8\}
\item
  top\_k \(\in\) \{50\} (keep fixed)
\end{itemize}

For each batch of synthetic tweets, I sample an additional 1,000 real
tweets and split the corpus into a training set (75\%) and an evaluation
set (25\%). I train an SVM classifier to discriminate between real and
synthetic tweets by attempting to predict if a tweet is real or
synthetic.\footnote{I use scikit-learn's SVM implementation (Pedregosa
  et al. 2011)}.

\begin{figure}[h]
\centering
\includegraphics{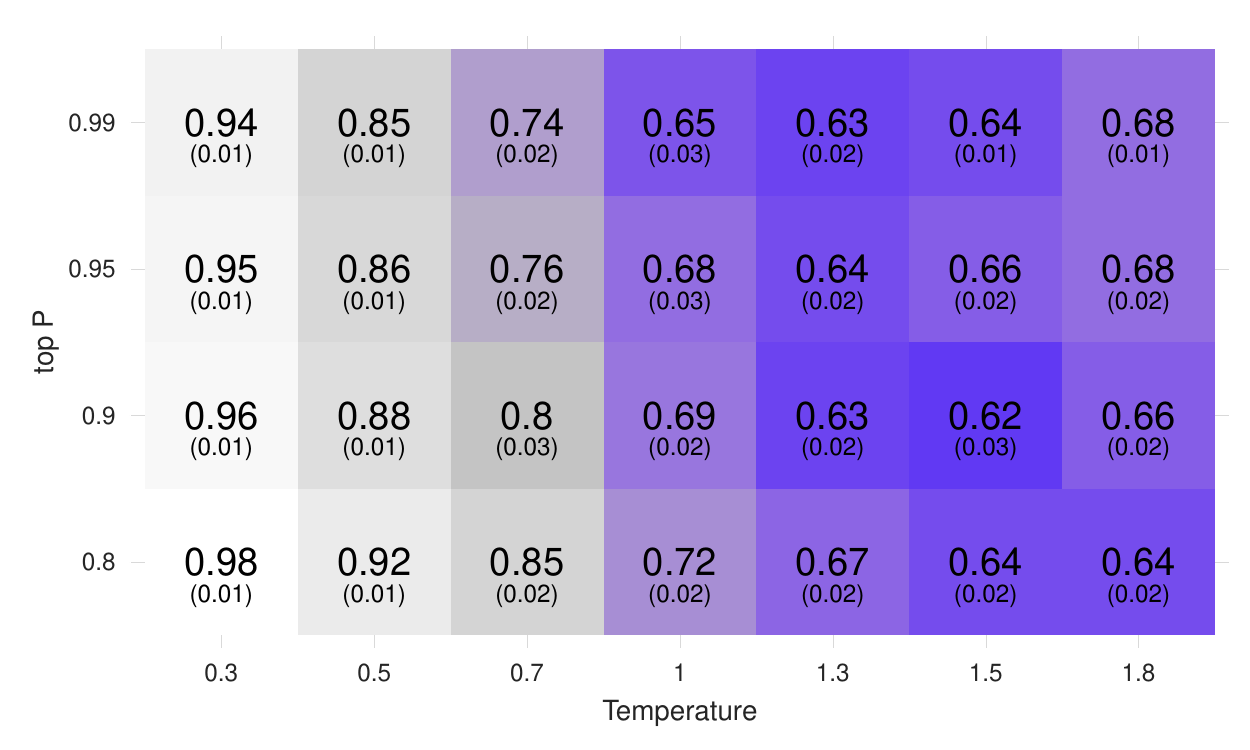}
\caption{Effect of GPT-2 generation hyperparameters on synthetic tweet quality. Accuracy is the mean out-of-sample accuracy of an SVM classifier trained to discriminate between real and synthetic tweets. Lower accuracy implies better synthetic tweet quality, as the model has a harder time distinguishing real and synthetic tweets. Numbers in parentheses report standard deviation across 10 runs (varying the sampled real tweets and the train/test split).}
\label{fig:tweet_gen_parameters}
\end{figure}

Figure \ref{fig:tweet_gen_parameters} shows the ability of a classifier
to distinguish between real tweets and the synthetic tweets generated
from each set of hyperparameters.

\hypertarget{sample-of-generated-tweets}{%
\subsubsection{Sample of generated
tweets}\label{sample-of-generated-tweets}}

The first ten tweets generated by a adapted/fine tuned GPT-2 model using
the optimal generation hyperparameters obtained from the process
described above: top\_p= 0.90, top\_k= 50, temperature= 1.5, training
epochs=3.

\begin{table}
\begin{tabular}{p{0.2cm}p{14cm}}
\hline
1. & \textsf{[SYNTH]} Yes, this is the USS Abraham Lincoln (DDG 71), which was in the Mediterranean a few weeks ago. And yes, you absolutely need to have a Go-Fund Ukraine account to donate money to charity organizations, including The Red Cross. \\
2. & \textsf{[SYNTH]}  An Osint Bunker article  detailing the recent Russian buildup near Kyiv \\
& You can click on the link to be taken to the actual map \\
3. & \textsf{[SYNTH]}   Russian military is getting massed in Belarus, most likely to set up new staging areas. \\
4.  & \textsf{[SYNTH]} The system is relatively good at engaging low/medium armored targets, like BTRs, MT-LBs, APCs and SPGs \\
5. & \textsf{[SYNTH]}  I think people got the wrong impression from today’s press conference, where Lukashenko said “I do not fear Western military threats but Russia is prepared to pay a heavy price for any military action. \\
6. & \textsf{[SYNTH]}  And as usual people are falling for the bait and trying to equate this to some sort of new high crime, especially when we see examples of this already in the news. \\
& We already saw it yesterday with  @Nrg8000 \\
7. & \textsf{[SYNTH]}  This is mostly because air defence is weak, and even non TB2s could get shot down. Only a very few aircraft flew today, with the majority of them from the western part of Ukraine. In the north of Ukraine a lack of TB2s has caused large losses. The Ukrainians are probably using the drones to spot artillery strikes. \\
8. & \textsf{[SYNTH]}  Tanks on the other side of the Irpin River \\
9. & \textsf{[SYNTH]}  Russian forces pushed back from Kharkiv tonight\\
& Kherson Oblast \\
10. & \textsf{[SYNTH]} It doesn't even have infrared sensors - only a SINCGARS system. This basically tells you what its main purpose is.
\end{tabular}

\caption{Synthetically generated tweets from a GPT-2 model adaptated/fine-tuned on 20,000
tweets reporting open source intelligence on the war in Ukraine. Due to
Twitter's restrictions on including actual tweets in published work, no comparison is provided for real
tweets. GPT-2 generation parameters: top\_p= 0.90, top\_k= 50, temperature= 1.5, fine tuning epochs=3}
\label{tab:generated_tweets}
\end{table}

\clearpage

\hypertarget{supplemental-information-b-headlines-for-prompting-events}{%
\subsection{\texorpdfstring{Supplemental Information B: Headlines for
prompting \textsf{ASSAULT}
events}{Supplemental Information B: Headlines for prompting  events}}\label{supplemental-information-b-headlines-for-prompting-events}}

\renewcommand{\thefigure}{B\arabic{figure}}
\renewcommand{\thetable}{B\arabic{table}}
\setcounter{figure}{0}
\setcounter{table}{0}

Synthetic news stories describing \textsf{ASSAULT} stories were prompted
by using around 40 headlines (see below). To increase the geographic and
stylistic diversity of the corpus, I followed the following process. For
each headline, I sampled five cities from the list below to generate a
dateline and varied the news source. For each unique (headline, city,
source) combination, I then generated five random stories.

Negative examples were generated from headlines prompting other event
types (protests, meetings, etc.), which are not shown for space reasons.

cities = \{``Abuja'', ``Kabul'', ``Belgrade'', ``Zagreb'', ``Khartoum'',
``Vienna'', ``Dhaka'', ``Brussels'', ``Minsk'', ``Kinshasa'',
``Beijing'', ``Bogota'', ``Sao Paulo'', ``Havana'', ``Berlin'',
``Prague'', ``Moscow'', ``Washington'', ``Cairo'', ``Jerusalem'',
``Delhi'', ``Tehran'', ``Rome'', ``Amman'', ``Beirut'', ``Tokyo'',
``Nairobi'', ``New York'', ``Panama City'', ``Oslo'', ``Damascus'',
``Bangkok'', ``Istanbul'', ``London'', ``Abu Dhabi''\}

sources = \{``Reuters'', ``AFP'', ``local sources''\}

\begin{verbatim}
"Activist held for ransom by rebel forces"
"Plane hijacked in suspected terrorist act"
"Police beat dozens of protestors"
"Syrian military tortured prisoners, new report states"
"Evidence of torture uncovered in government prison"
"Two prisoners put to death"
"Execution carried out in Iranian prison"
"Two women raped in capital"
"New reports of sexual violence in ongoing conflict"
"Al Qaeda leader killed in targeted drone strike"
"Iranian scientist assassinated"
"Russian dissident killed in apparent assassination"
"Two civilians hacked to death with machetes in rural area"
"Buildings destroyed in shelling"
"Israeli bulldozers clear houses"
"Four shot in ongoing gun battle"
"Police fire on demonstrating crowd"
"Truck bomb kills three soldiers"
"Explosions rock city"
"Landmine kills three"
"Bomb detonates in downtown capital"
"Dozens killed in suicide bombing"
"Suicide bomber kills three"
"Man detonates explosive vest at checkpoint"
"US tanks and armored vehicles begin assault in Iraqi town"
"Heavy artillery shelling continues"
"Police disperse protest with water cannons and tear gas"
"Police fire weapons in the air to disperse mob"
"Indian police use lathi charge to break up protest"
"Hundreds expelled from homes in ethnic cleansing"
"Ethnic cleansing ongoing in conflict"
"Serbian forces expel Bosnians from villages in cleansing operation"
"Civilians slaughtered in massacre"
"Syrian air force uses chemical weapons against civilians"
"Four killed in sarin gas attack"
"Anthrax attack infects three"
"Four killed in air strike"
"War planes pummel rebel positions"
"Allied aircraft enforce no-fly-zone, shooting down Iraqi fighter plane"
"Air Force UAV destroys enemy targets"
"Drone strikes increase as conflict intensifies"
"Man hacked to death with machete"
"Angry mob throws rocks and bottles"
"Local opposition leader beaten with baseball bat"
"Terrorist group releases poison gas, killing three"}
\end{verbatim}

Thus, an example of a complete prompt would be ``Suicide bomber kills
three HAVANA (local sources) --''.

\clearpage

\renewcommand{\thefigure}{C\arabic{figure}}
\renewcommand{\thetable}{C\arabic{table}}
\setcounter{figure}{0}
\setcounter{table}{0}

\hypertarget{supplemental-information-section-c-populism-prompts-and-validation}{%
\subsection{Supplemental Information Section C: Populism Prompts and
Validation}\label{supplemental-information-section-c-populism-prompts-and-validation}}

Table \ref{tab:cmp_prompts} shows an example non-populist prompt using
the Manifesto Project codebook. Table
\ref{tab:non_populist_hand_prompts} below show the prompts used to
generate non-populist text. Table \ref{tab:ukip_random} shows 10
randomly selected sentences from United Kingdom Independence Party
manifestos and their predicted populism scores.

\begin{table}[h]
\renewcommand{\arraystretch}{1.5}
\begin{tabular}{p{0.3\textwidth}p{0.7\textwidth}}
Economic Orthodoxy (414) & A political party is calling for economic orthodoxy, the need for economically healthy government policy making. May include calls for:\newline • Reduction of budget deficits;\newline• Retrenchment in crisis;\newline • Thrift and savings in the face of economic hardship;\newline • Support for traditional economic institutions such as stock market and banking system;\newline • Support for strong currency.\newline Write a list of 10 statements that this party in \{country\} might make (in the \{language\} language) in its party platform: \\
Welfare State Expansion (504.0) & A political party supports welfare state expansion. Favourable mentions of need to introduce, maintain or expand any public social service or social security scheme. This includes, for example, government funding of:\newline
• Health care;\newline
• child care;\newline
• Elder care and pensions;\newline
• Social housing.\newline
Note: This category does NOT include education. \newline
Write a list of 10 statements that this party in \{country\} might make (in the \{language\} language) in its party platform: \\
\end{tabular}
\caption{Example Comparative Manifesto Project prompts for non-populist text. The ten most
common codes are 414, 201.0, 416.2, 504.0, 403.0, 703.0, 304.0, 402, 705.0,
502.0. See Volkens et al. (2021) for details.}
\label{tab:cmp_prompts}
\end{table}

\begin{table}[h]
\renewcommand{\arraystretch}{1.5}
\begin{tabular}{p{0.8\textwidth}}
  \hrule
    ``A political party supports existing political institutions, multiculturalism, globalization, and respect for the existing political process. It supports equality for all people and welcomes immigrants.\newline Write ten statements that this party in \{country\} might make (in \{language\}):"\\
    ``A party in \{country\} believes that taxes are too high. It wants to reduce taxes and cut government spending. Write 10 statements that \{adjective\} party might make in the style of a political manifesto (in the \{lang\} language):"\\
    ``A party in \{country\} wants to spend more money on schools, housing, and the military. Write 10 statements that \{adjective\} party might make in the style of a political manifesto (in the \{lang\} language):"\\
    ``A party in \{country\} believes that the government should spend more money on healthcare, education, and infrastructure. Write 10 statements that \{adjective\} party might make n the style of a political manifesto (in the \{lang\} language):"\\
    ``A party in \{country\} wants to raise the minimum wage. Write 10 statements that \{adjective\} party might make in the style of a political manifesto (in the \{lang\} language):"\\
    ``A party in \{country\} is calling for greater support for teachers and police. Write 10 statements that \{adjective\} party might make in the style of a political manifesto (in the \{lang\} language):"\\
    ``A party in \{country\} is criticizing the foreign policy of its opposing party. Write 10 statements that \{adjective\} party might make in the style of a political manifesto (in the \{lang\} language):"\\
    ``A party in \{country\} is criticizing populism as a threat to \{country\}. Write 10 statements that \{adjective\} party might make in the style of a political manifesto (in the \{lang\} language):"\\
    ``A party in \{country\} is criticizing the high rate of unemployment. Write 10 statements that \{adjective\} party might make in the style of a political manifesto (in the \{lang\} language):"\\
    ``A party in \{country\} believes the country is on the wrong path. Write 10 statements that \{adjective\} party might make in the style of a political manifesto (in the \{lang\} language):"\\
    ``A party in \{country\} is criticizing the party is is campaigning against. Write 10 statements that \{adjective\} party might make in the style of a political manifesto (in the \{lang\} language):"
      \hrule
\end{tabular}
\caption{Hand-written prompts used to generate non-populist text. ``adjective'' refers to the country adjective such as ``Swiss" or ``Dutch".}
\label{tab:non_populist_hand_prompts}
\end{table}

\begin{table}[h]
\renewcommand{\arraystretch}{1.5}
\centering
\begin{tabular}{p{0.8\textwidth}p{0.1\textwidth}}
  \hline
  Sentence & Score\\ 
  \hline
 As a minimum, we will seek continued access on free-trade terms to the EU’s single market. & 0.05 \\ 
 A GRAMMAR SCHOOL IN EVERY TOWN The state education system of grammar, secondary modern and technical schools was designed to make a high standard of education available to all, irrespective of social background. & 0.06 \\ 
 Climate Change Act, the most expensive piece of legislation in  history. & 0.06 \\ 
 BRITISH CULTURE & 0.05 \\ 
 3. & 0.09 \\ 
 In the longer term, we will aim to restore the personal allowance to those earning over £100,000 and make 40 per cent the top rate of tax for all, as it used to be. & 0.06 \\ 
 in particular for children, the elderly and people on low incomes. & 0.04 \\ 
 We will also exempt foodbanks and charity shops from charges imposed by local authorities to dispose of unwanted food waste and other goods. & 0.05 \\ 
 UKIP will restore Britain’s armed forces to their rightful place among the most professional, flexible and effective fighting forces in the world, and we will sign a new military covenant with our brave heroes. & 0.12 \\ 
 Prisoners are encouraged to deal with drug addiction problems during their incarceration, and we do them no favours by not taking a tough line. & 0.05 \\ 
   \hline
\end{tabular}
\caption{Ten randomly selected UKIP manifesto sentences and their predicted populist scores.}
\label{tab:ukip_random}
\end{table}

\begin{table}[h]
\renewcommand{\arraystretch}{1.5}
\begin{tabular}{p{0.9\textwidth}}
\hline
``Far dimagrire lo Stato, i cittadini danno già troppo"\\
\hspace{1em}[translated] Major slimming down of the State: citizens already give too much\\
``Incremento della lotta per la legalità, per il contrasto ai fenomeni della immigrazione clandestina"\\
\hspace{1em}[translated] "Increase in the fight for legality, to contrast the phenomena of illegal immigration"\\
``Nuove azioni per favorire la concorrenza nel settore energetico e contrastare gli oligopoli"\\
\hspace{1em}[translated] "New actions to encourage competition in the energy sector and counter oligopolies"\\
``Razionalizzare la distribuzione territoriale degli istituti e degli insegnamenti universitari"\\
\hspace{1em}[translated] "Rationalize the territorial distribution of institutes and university teaching"\\
``Eventuali salvataggi bancari devono essere solo a tutela dei risparmiatori e non degli azionisti di controllo"\\
\hspace{1em}[translated] "Any bank bailouts must only protect savers and not controlling shareholders"\\
``Votare il dimezzamento degli emolumenti dei parlamentari."\\
\hspace{1em}[translated] "Vote to halve the salaries of parliamentarians."\\
``Elezione diretta e popolare del Presidente della Repubblica"\\
\hspace{1em}[translated] Direct and popular election of the President of the Republic\\
``Dimezzamentodei costi della politica. Abolire il finanziamento pubblico dei partiti (nessun fondo pubblico ai partiti)"\\
\hspace{2em}[translated] "Halving of the costs of politics Abolishing public funding of parties (no public funding to parties)"\\
``Più Europa dei Popoli, meno euro-burocrazia"\\
\hspace{2em}[translated] "More Europe of Peoples, less Euro-bureaucracy"\\
``La difesa e il sostegno alla famiglia, comunità naturale fondata sul matrimonio tra uomo e donna"\\
\hspace{2em} [translated] "The defense and support of the family, a natural community founded on marriage between a man and a woman" \\
\hline
\end{tabular}
\caption{Hand validation of populism in Italy's Northern League 2013 manifesto, showing the sentences that appear most populist in a hand coding of the document. The populism classifier identifies no sentences with a predicted populism score above 0.5, while PopuList codes the party-election as populist. While some of the sentences contain slightly populist statements, none appear to be overtly anti-elite and pro-common person.}
\label{tab:northern_league_hand}
\end{table}

\end{document}